# Graph Neural Network Framework for Sentiment Analysis Using Syntactic Feature

Linxiao Wu
Columbia University
New York, USA

Yuanshuai Luo
Southwest Jiaotong University
Chengdu, China

Binrong Zhu
San Francisco State University
San Francisco, USA

Guiran Liu
San Francisco State University
San Francisco, USA

Rui Wang
Carnegie Mellon University
Pittsburgh, USA

Qian Yu*
Trine University
Detroit, USA

*Abstract*—Amidst the swift evolution of social media platforms and e-commerce ecosystems, the domain of opinion mining has surged as a pivotal area of exploration within natural language processing. A specialized segment within this field focuses on extracting nuanced evaluations tied to particular elements within textual contexts. This research advances a composite framework that amalgamates the positional cues of topical descriptors. The proposed system converts syntactic structures into a matrix format, leveraging convolutions and attention mechanisms within a graph to distill salient characteristics. Incorporating the positional relevance of descriptors relative to lexical items enhances the sequential integrity of the input. Trials have substantiated that this integrated graph-centric scheme markedly elevates the efficacy of evaluative categorization, showcasing preeminence across a spectrum of established reference datasets.

*Keywords-Graph neural network, sentiment analysis, syntactic analysis tree*

## I. INTRODUCTION

With the accelerating proliferation of social media networks and e-commerce platforms, discerning public sentiment has emerged as a critical focus within the realm of computational linguistics [1]. A specialized subset of this domain involves pinpointing the nuanced sentiments associated with specific topics within text segments, encompassing sentiments ranging from favorable to unfavorable or neutral stances [2]. Practical instances of this methodology are prevalent in analytics on medical dignosis[3-5], and insights derived from financial stakeholders' communications[6]. This targeted sentiment extraction holds substantial academic and commercial significance [7].

The deployment of deep learning paradigms has catalyzed advancements in this niche [8]. Traditionally, addressing this challenge involved transforming textual inputs into lower-dimensional vectors via embedding models before feeding them into neural architectures for training. However, this direct approach often overlooks the influence of word sequence on the model's efficacy. Moreover, incorporating syntactic-level information could enhance the model's discernment of sentiment toward specific topics. Utilizing the hierarchical structure of sentences, as depicted in syntactic parse trees, can bolster the extraction of meaningful syntactic features. Grounded in these considerations, this study employs graph neural networks to address these issues, contributing to the field in three primary ways.

(1) An integrated graph neural network framework that incorporates the positional context of focal terms is introduced. The structural relationships within the input text are mapped into a matrix form, facilitating feature extraction through the application of graph-based convolution and attention mechanisms.

(2) To maintain the sequential arrangement of lexical units within the input, the relative proximity of focal terms is utilized as a positional attribute, ensuring that the inherent order is preserved as a feature.

(3) Ultimately, these feature vectors are channeled into a retrieval-oriented attention component, aiding a SoftMax classifier in deriving the conclusive classification outcome. Evaluations conducted on three canonical datasets affirm the model's validity and underscore the significance of each component in addressing the challenges posed by text sequence processing and the utilization of graph neural networks in achieving superior results.

## II. RELATED WORK

The domain of sentiment analysis, particularly using deep learning and syntactic features, has been significantly advanced by various methodologies and models in recent years. These developments are essential for the understanding of how the intersection of graph neural networks (GNNs) and sentiment analysis can lead to more precise and contextually aware results.

Recent work on text classification using GNNs, such as Gao et al.'s [9], showcases the effectiveness of graph-based models in capturing syntactic dependencies, which is crucial in sentiment analysis where sentence structure plays a key role. Their study introduces a text classification optimization algorithm based on GNNs, which aligns with the proposed approach of syntactic feature integration in sentiment analysis.

Additionally, Yang et al. [10] explored emotional analysis using large language models, which are inherently designed to handle nuanced language processing tasks. This paper, while focusing on emotional analysis, contributes significantly to understanding how deep learning techniques can be harnessed for sentiment extraction, a close cousin to opinion mining, by leveraging advanced language models. Embedding techniques such as those in Cheng et al.'s [11] work, which involves ELMo word embeddings and multimodal transformers, further push the boundaries of sentiment analysis. Their research emphasizes the need for robust contextual embedding models that capture syntactic and semantic features, resonating with the need for high-dimensional feature representation in the proposed GNN framework.

In terms of optimization strategies, Ma et al. [12] delved into gradient descent optimization, which is vital for training deep neural networks, including GNNs. The methods discussed provide a solid foundation for improving model training in complex neural network architectures such as the one proposed in this research. Furthermore, Wang et al. [13] have contributed to the understanding of syntactic feature extraction with their research on enhancing convolutional neural networks using numerical difference methods. Their approach is highly relevant to improving the efficiency of GNNs in dealing with syntactic structure extraction, a key focus of this study. Several studies, though not directly focused on sentiment analysis, offer crucial insights into deep learning and graph-based techniques. For instance, Yang et al. [14] presented advancements in dynamic hypergraph prediction, which shares conceptual similarities with the graph-based approach for modeling relationships between syntactic elements in text. Likewise, Gu et al. [15] researched spatio-temporal aggregation in graph models, which informs the methodology of integrating temporal and positional features within GNN frameworks for sentiment analysis.

Finally, recent studies on named entity recognition [16] and risk assessment in financial markets [17] also illustrate the broad applicability of deep learning and graph-based methods in diverse fields. These works underscore the versatility of GNNs and their potential for cross-domain adaptation, strengthening the argument for their use in sentiment analysis.

III. METHOD

Using graph neural networks (GNNs) for modeling often relies on a single-layered approach, which may limit the exploration of the intricate relationships between focal terms and their surrounding context[18-20]. This method can miss out on capturing the syntactic constraints inherent in sentences. Additionally, simply applying a model to the sentence without considering the positional information can lead to misinterpretation, as it might rely on irrelevant context for sentiment determination. For example, attention-based models are prone to highlighting words that are not necessarily relevant to the sentiment being analyzed, thereby missing key syntactic dependencies. Moreover, from observations in aspect-based sentiment analysis datasets, it becomes evident that context words located nearer to the focal term within a sentence tend to carry more weight in determining sentiment than those positioned further away. This phenomenon is clearly demonstrated in Figure 1, showing the importance of proximity in sentiment classification.

To address these limitations, a more sophisticated approach is required that not only takes into account the syntactic structure and positional information but also leverages the strengths of GNNs to accurately capture the nuanced interactions between different parts of the sentence. This ensures that the model can effectively distinguish between relevant and irrelevant context, thereby improving the accuracy of sentiment analysis.

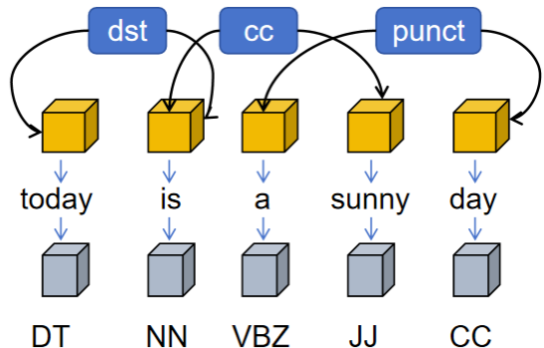

Figure 1 Parsing tree

First, a syntactic parser is employed to transform the given sentence into a syntactic parse tree. This tree is subsequently translated into an adjacency matrix, which serves as the foundation for constructing various types of graph neural networks. These networks are designed to extract the syntactic dependency information among the context words comprehensively.

Secondly, to emphasize the positional sequence of context words relative to the focal terms within the sentence, a method based on the relative distance to these focal terms is developed. This relative distance is incorporated as positional sequence information. The positional sequence information is then combined with the sentence context and embedded as the input feature vector for the model. Following this, a BiLSTM-Attention layer processes the input to generate a hidden vector representation of the text.

Finally, this hidden vector representation, along with the output features from the graph neural network component, is fed into an Attention Weight module. This module assigns attention weights to the context words, aiding in the final classification outcome through a SoftMax classifier. The architecture of this network is illustrated in Figure 2.

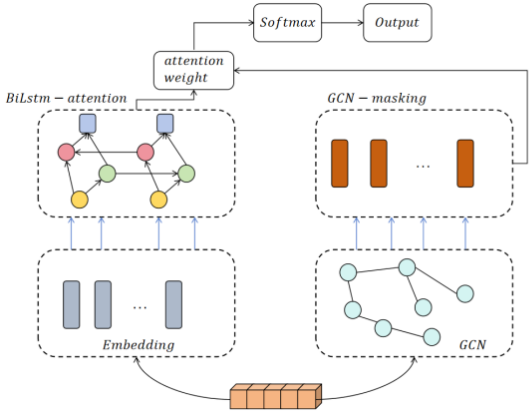

Figure 2 network architecture

The graph attention network (GAT) consists of multiple layers designed to aggregate the weights of neighboring nodes effectively[21]. This process starts by incorporating updates from the syntactic parse tree nodes into a single-layer GAT, where each node's update is computed as follows:

$$ei_j = (Concat(Wh_i^g, Wh_i^g))a$$

$$a_{ij} = \frac{\exp(Leaky\,\mathrm{Re}\,LU(e_{ij}))}{\sum_{j \in N_i} \exp(Leaky\,\mathrm{Re}\,Lu(ei_j))}$$

$$h_i = \mathrm{Re}\,LU(\sum_{j \in N_i} a_{ij}^W h_i^g)$$

Where $h_i$ and $h_j$ denote the feature vectors of nodes i and node j in $H$, with W representing the trainable matrix for transforming feature vectors. The hidden state output $H = \{h_1, h_2, ... h_{n+m}^a\} \in R^{2d_h*(n+m)}$ is obtained. After the nodes are updated, the resulting vectors are utilized to derive the vectors of neighboring nodes with varying contribution weights. Graph convolutional neural networks (GCN) excel in handling graph data rich in dependencies. By aggregating the obtained hidden vectors through GCN, the system can effectively refine the scope of aggregation around focal terms and sentiment-expressing words, making the combination more efficient. The process of aggregating and updating the nodes can be described as follows:

$$h_i = \mathrm{Re}\,LU(\sum_{j \in N_i} A_{ij} W^{(l)} h_i^a + b^l)$$

## IV. EXPERIMENT

### A. Experimental setup

In our experimental framework, we have systematically configured all parameters to ensure robustness and replicability. We initiate by employing pre-trained GloVe embeddings for initializing word vectors, ensuring that both context and aspect words have an embedding dimensionality of 300. Position vectors are distinctly configured with a dimensionality of 100. The weight matrix for the final fully connected layer is initialized at random, adhering to a normal distribution characterized by a mean of 0 and a standard deviation of 1. For the optimization process, the Adam optimizer is selected for its efficiency in handling sparse gradients on noisy problems[22]. We set the initial learning rate at 0.001, coupled with an L2 regularization parameter finely tuned to 0.00001. This configuration is designed to balance learning efficiency with the need to prevent overfitting, thus enhancing the model's performance on unseen data. This methodical setup forms the backbone of our experimental architecture, aiming to foster reproducibility and reliability in our results.

To assess the model's performance, we utilize two primary evaluation metrics: accuracy (Acc) and macro F1 score. These criteria provide a comprehensive measure of the model's effectiveness across different aspects of the classification task. The accuracy metric evaluates the proportion of correctly classified instances, whereas the macro F1 score offers a balanced view of precision and recall, averaged across all classes. This setup ensures a thorough and balanced evaluation of the model's capabilities.

### B. Datasets

Three separate datasets were employed in the experiments, and the results validated the effectiveness of the proposed HM-GNN model. The datasets included the Laptop and Restaurant categories from the SemEval2014 competition, as well as a Twitter dataset from ACL2014. To maintain consistency, instances with conflicting sentiment polarity labels were excluded from the SemEval2014 datasets. Each dataset consisted of sentences, the specific aspect terms within them, and the corresponding sentiment polarities for those terms. These datasets were processed using the Linked Data approach, which integrates multiple data formats, a method essential for academic research[23]. This structured technique enhances data cross-referencing, promoting interoperability among different datasets. Such capabilities are particularly advantageous in fields like machine learning and artificial intelligence, where the quality of data is critical for effective model training and ensuring accurate results.

The experimental setup involved analyzing the textual content, identifying the focal aspects within each sentence, and assessing the sentiment polarity related to these aspects. By removing examples with inconsistent sentiment labels, the datasets were refined to provide a clearer basis for evaluating the model's performance. This methodology was pivotal in structuring the datasets to rigorously evaluate the model's effectiveness in accurately classifying sentiment polarities. By meticulously preparing the data, we ensured that each dataset was optimally configured for comprehensive testing, thereby facilitating a thorough assessment of the model's performance in different scenarios. This structured approach not only enhances the reliability of the results but also provides a solid foundation for comparing the predictive capabilities of the model under controlled conditions. Through this rigorous preparation, we aim to demonstrate the robustness and precision of the model in discerning various sentiment polarities across diverse textual datasets.

*C. Experimental Results*

To thoroughly evaluate the performance of the HM-GNN model, we engaged in a detailed comparative analysis against a diverse array of baseline models. This lineup included some well-known models, representing distinct methodologies within the field. By incorporating such a wide spectrum of approaches, our analysis provided a robust benchmark for assessing the HM-GNN model's capabilities. This comprehensive evaluation strategy allowed us to discern the strengths and weaknesses of HM-GNN in contrast to established methods, thereby offering insights into its comparative effectiveness and potential areas for further enhancement. This benchmarking is crucial for validating the novel contributions of the HM-GNN model and establishing its practical utility in sentiment analysis tasks.

The evaluation aimed to highlight the strengths and weaknesses of HM-GNN relative to these established models. By comparing the results, we were able to determine how effectively the HM-GNN model performs in terms of accuracy and efficiency, offering insights into its potential advantages and areas for improvement. This comparative study ensured a thorough understanding of the model's performance across different methodologies.

Table 1    Model experimental results in Twitter dataset

| Model | Acc | F1 |
|---|---|---|
| LSTM | 65.51 | 66.11 |
| TD-LSTM | 66.32 | 66.56 |
| TNet-LF | 67.78 | 67.63 |
| AEN | 68.29 | 68.42 |
| ASGCN | 70.13 | 69.23 |
| DGEDRT | 71.33 | 70.99 |
| RPAEN | 72.47 | 71.65 |
| BATX | 73.97 | 71.73 |
| Ours | 74.36 | 72.79 |

Table 2    Model experimental results in Laptop dataset

| Model | Acc | F1 |
|---|---|---|
| LSTM | 66.58 | 60.11 |
| TD-LSTM | 70.97 | 63.71 |
| TNet-LF | 71.46 | 64.56 |
| AEN | 73.23 | 65.23 |
| ASGCN | 75.42 | 66.73 |
| DGEDRT | 76.06 | 67.49 |
| RPAEN | 76.12 | 70.32 |
| BATX | 76.38 | 71.56 |
| Ours | 77.43 | 73.44 |

Table 3    Model experimental results in Restaurant dataset

| Model | Acc | F1 |
|---|---|---|
| LSTM | 74.35 | 63.08 |
| TD-LSTM | 75.32 | 66.23 |
| TNet-LF | 76.77 | 68.73 |
| AEN | 77.13 | 69.52 |
| ASGCN | 78.28 | 70.23 |
| DGEDRT | 79.33 | 70.99 |
| RPAEN | 79.47 | 71.65 |
| BATX | 80.97 | 72.73 |
| Ours | 81.87 | 73.13 |

In order to further and more intuitively show the superiority of our algorithm, we draw a bar chart based on the Twitter dataset, as shown in Figure 3

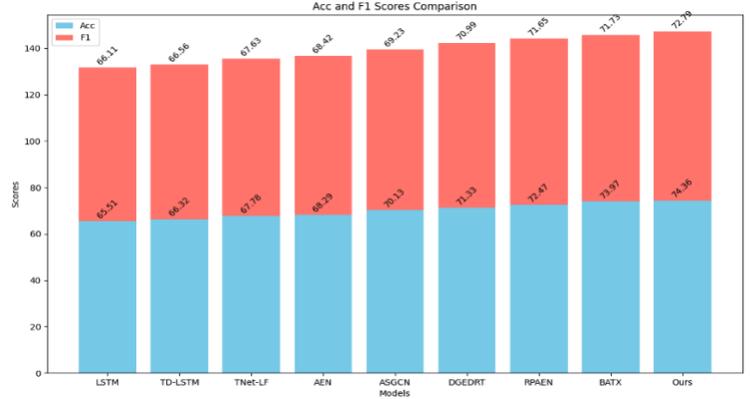

Figure 3 Twitter dataset results in bar chart

As can be seen intuitively from Figure 3, whether it is the F1 value or the ACC value, our results are better than the previous algorithm performance and have reached the state of the art.

V. CONCLUSION

This paper introduces a novel hybrid graph neural network model that leverages the syntactic structure of input sentences. The syntactic parse tree is transformed into an adjacency matrix, which serves as the basis for feature extraction using a combination of graph convolutional neural networks (GCNs) and graph attention networks (GANs). To maintain the sequential information of words within the input, the model incorporates the relative distance from aspect words as a positional feature.

The feature vectors derived from this process are then passed through a retrieval-style attention module, enabling a SoftMax classifier to produce the final classification outcome. Validation conducted on three established datasets not only underscores the logical design of the model but also highlights the impact of each component in addressing the complexities of text sequence processing and the benefits of using graph neural networks.

The experimental results demonstrate that the proposed hybrid model surpasses several existing baseline models in terms of accuracy and F1 score, thereby confirming its effectiveness and superiority in the domain of sentiment analysis. This comprehensive evaluation provides evidence of the model's robust performance and its potential for advancing the field.